\DocumentMetadata{lang=en,pdfstandard=UA-2, pdfstandard=a-4f}
\documentclass[11pt]{article}

\usepackage[preprint]{acl}

\usepackage{microtype}

\usepackage[T1]{fontenc}
\usepackage[utf8]{inputenc}
\usepackage{newtxtext,newtxmath}
\usepackage{inconsolata}

\usepackage{graphicx}

\usepackage{float}

\usepackage{pgfplots}
\pgfplotsset{compat=1.18}
\usepgfplotslibrary{groupplots}
\usepgfplotslibrary{colormaps}

\usepackage{booktabs}
\usepackage{tabularx}

\title{Examining the Limits of Word2Vec with Toki Pona}

\author{Daniel Zhenhan Huang \\
  Georgia Institute of Technology \\
  Atlanta, GA \\
  \texttt{dzh@gatech.edu} \\\And
  Hongchen Wu \\
  Georgia Institute of Technology \\
  Atlanta, GA \\
\texttt{hwu480@gatech.edu} \\}

\begin{document}
\maketitle
\begin{abstract}
  Word2Vec's effectiveness at generating semantic embeddings has been widely validated, yet it has been tested almost exclusively on languages with large vocabulary inventories. This study examines whether Word2Vec can successfully capture semantic relationships within an extremely reduced vocabulary using data from Toki Pona, a constructed language with approximately 130 words. We sourced 1.4 million sentences (7.95 million tokens) from the Toki Pona community for training. Approximately 23\% of sentences in the corpus contain non-Toki Pona tokens such as named entities, loanwords, and neologisms. To investigate whether this linguistic noise enhances or hinders performance---a topic rarely addressed in word embedding literature---we trained two distinct models: one retaining these incidental tokens and another filtering them out completely. Evaluation was conducted using quantitative methods measuring word proximity to semantic category centroids, automated silhouette scores via agglomerative clustering, and qualitative analysis utilizing representational similarity matrices compared against English. The results indicate that while sparse, non-core tokens do not affect the relative structure of the learned embeddings, they actually draw similar words closer together in the vector space. Importantly, Word2Vec's effectiveness depends more on distributional patterns than lexicon size even at this extreme lower bound.
\end{abstract}

\section{Introduction}

Word embeddings are a foundational tool in natural language processing, capturing words' semantic relations from their distributional patterns in text \citep{mikolov2013efficientestimationwordrepresentations}. Their effectiveness has been repeatedly validated, but almost entirely on natural languages whose vocabulary size ranges from tens of thousands to hundreds of thousands of words. What happens when we push Word2Vec to an extreme edge case, significantly reducing its vocabulary's size?

Toki Pona is a lexically minimal constructed language created by Sonja Lang in 2001, designed for explicit and deliberate expression with a vocabulary of approximately 130 core words \citep{lang2014, asi2025}. As such, words in Toki Pona are often broad and polysemous: \textit{lipu}\footnote{In Toki Pona orthography, all core vocabulary is written in lowercase regardless of sentence position; only proper names, which are adapted to Toki Pona phonotactics and preceded by a noun, are capitalized. We follow this convention when citing Toki Pona forms. The English name of the language is capitalized per English convention.} refers to paper or flexible sheets, but also extends abstractly to documents like websites and blogs; \textit{tomo} covers physical enclosures like rooms, but also vehicles. The minimization of content words creates an environment radically different from those for which Word2Vec was designed.

Distributional models like Word2Vec rely on co-occurrence profiles between pairs of words to distinguish words' meanings. With a vocabulary of only 130 words, most words will co-occur with others at a much higher frequency, potentially making these profiles less discriminative. This raises a question about Word2Vec's efficacy: does an effective model require a large vocabulary with fine lexical distinctions to recover meaning, or can meaningful structure emerge even in Toki Pona's minimalist vocabulary?

A complicating factor is that approximately 23\% of sentences in the gathered corpus contain non-Toki Pona tokens---proper names, loanwords, and neologisms that fall outside the language's 130-word core vocabulary. While these tokens reflect natural use, their impact on embedding quality is unclear; they may provide additional context that improves the quality of semantic relationships, or they may obscure the structure of the core vocabulary. As such, we train two Word2Vec models on a corpus of 1.4 million Toki Pona sentences with and without these non-Toki Pona tokens and evaluate the resulting embeddings to understand what the language's nature reveals about the capabilities of word embedding models.

We further compare the models against an independently trained English model using representational similarity analysis. Consistent patterns between semantic categories across languages would provide convergent validity and indicate that the models learn real semantic structure, not just artifacts of the corpus or of our category definitions.

\section{Methodology}

\subsection{Data}

Despite its minimalist vocabulary, Toki Pona enjoys a speaking community of thousands \cite{meulen2021}. Because the language gained popularity primarily through the internet, the vast majority of its use is online. Its largest speaking community, \textit{ma pona pi toki pona} (lit. `a good place for Toki Pona'), is a Discord server with over 18,000 members \cite{mapona2025}.

The corpus used in this paper consists of conversational data collected from public channels in this community, spanning from early 2016 through November 2025. After filtering to retain only Toki Pona sentences with \citeposs{danielson2025} \textit{sona toki} library and removing single-token sentences, the corpus contains a total of 1.42 million sentences with 7.95 million tokens.

While \citeposs{danielson2025} method reliably determines whether a sentence can be considered to be in Toki Pona using heuristics such as Toki Pona word frequency and phonotactic conformity, many sentences still contain incidental non-Toki Pona words such as named entities (e.g., \textit{Sonja}, the language's creator, and \textit{Inli}, the Toki Pona adaptation of \textit{English}) and other loanwords. These non-Toki Pona tokens may serve as ``noise'', obscuring the semantic relationships between core vocabulary items, or they may provide additional context that ultimately strengthens the embeddings' effectiveness. To better understand the impact of these additional tokens on the semantic structure of the model, a subset of the corpus is created with all sentences containing non-Toki Pona tokens removed, totaling 1.10 million sentences and 5.90 million tokens. We refer to these as the \textbf{Full Corpus model} and \textbf{Pure Toki Pona model} respectively.

\begin{figure}[t]
  \centering
  \begin{tikzpicture}
    \begin{axis}[
        xbar,
        title={\textbf{A. Total Token Occurrences}},
        width=0.9\columnwidth,
        height=4.5cm,
        symbolic y coords={Abbreviations, Miscellaneous, Unadapted, Nonstandard, Named Entities},
        ytick=data,
        nodes near coords,
        nodes near coords style={
          /pgf/number format/fixed,
          /pgf/number format/precision=0,
          font=\scriptsize,
          anchor=west
        },
        tick label style={font=\scriptsize},
        label style={font=\small},
        title style={font=\small},
        scaled x ticks = false,
        xticklabel style={/pgf/number format/fixed},
        enlarge x limits={upper, value=0.3},
        xmin=0,
        xlabel={Count},
        ymajorgrids=false,
        xmajorgrids=true,
        grid style=dashed,
        bar width=12pt,
      ]
      \addplot coordinates {
        (3386,Abbreviations)
        (19248,Miscellaneous)
        (38389,Unadapted)
        (55062,Nonstandard)
        (113663,Named Entities)
      };
    \end{axis}
  \end{tikzpicture}%
  \hfill
  \begin{tikzpicture}
    \begin{axis}[
        xbar,
        title={\textbf{B. Unique Token Types}},
        width=0.9\columnwidth,
        height=4.5cm,
        symbolic y coords={Abbreviations, Miscellaneous, Unadapted, Nonstandard, Named Entities},
        ytick=data,
        nodes near coords,
        nodes near coords style={font=\scriptsize, anchor=west},
        tick label style={font=\scriptsize},
        label style={font=\small},
        title style={font=\small},
        enlarge x limits={upper, value=0.3},
        xmin=0,
        xlabel={Unique Types},
        xmajorgrids=true,
        grid style=dashed,
        bar width=12pt,
      ]
      \addplot coordinates {
        (15,Abbreviations)
        (69,Miscellaneous)
        (212,Unadapted)
        (103,Nonstandard)
        (519,Named Entities)
      };
    \end{axis}
  \end{tikzpicture}
  \caption{Distribution of non-Toki Pona tokens. Named entities adapted to Toki Pona phonotactics make up the largest fraction of the volume of removed words. Though nonstandard words make up a large volume of tokens, unadapted loanwords contribute a higher number of unique types.}
  \label{fig:omitted-tokens}
\end{figure}
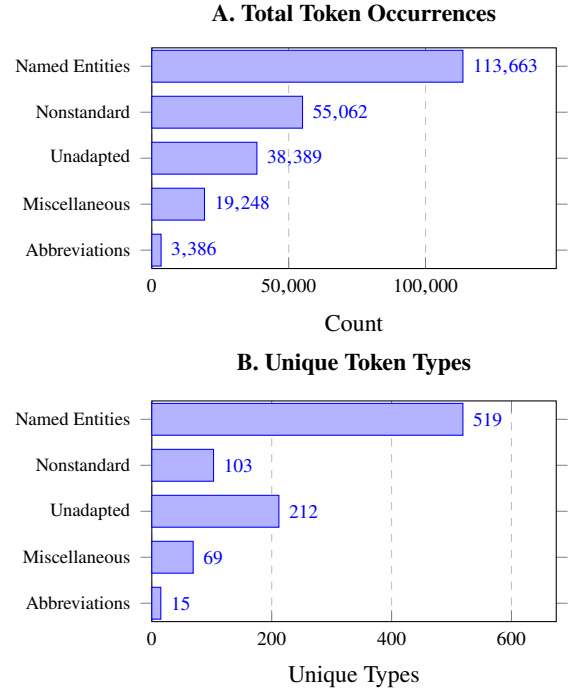

We categorized token types present in the Full Corpus but not the Pure Toki Pona corpus with at least 50 occurrences, as shown in Figure~\ref{fig:omitted-tokens}. The most frequent omissions were those of \textit{named entities}, i.e., proper names adapted to fit Toki Pona's phonotactics like \textit{Sonja} and \textit{Inli}, making up over 110,000 tokens. Nonstandard words (e.g., \textit{lanpan} `to take, steal') and neologisms (e.g., \textit{eliki} `adversity; bittersweet') were the second largest category by usage, with over 55,000 occurring tokens. Unadapted words, i.e., direct loans such as \textit{Discord} from non-Toki Pona languages without adaptation, appeared less frequently than nonstandard words but represented a larger number of unique types, suggesting that code-switching, while infrequent, spans a wide variety of terms. In the following sections, we will examine whether these words provide additional useful information or simply introduce ``noise'' into the Full Corpus model when generating word embeddings.

\subsection{Training}

Two models are trained from scratch---one on the Full Corpus and the other on the Pure Toki Pona corpus.

The models are trained with skip-gram with a window size of 5 over 10 epochs. Skip-gram was chosen for its stronger performance in semantic tasks \cite{mikolov2013efficientestimationwordrepresentations}. A major consideration during training is the vector dimensionality hyperparameter; typical models for natural languages use between 100 and 300, but these are for vocabularies of hundreds of thousands of words. \citet{yin2018dimensionality} develop methods to determine the optimal dimensionality based on a lexicon size, provided corpus, and training algorithm; this was used on the corpus and vocabulary, yielding an optimal size of 24 dimensions. Trained models (see Appendix~\ref{sec:code}) are available along with the publication of this paper.

\subsection{Model Evaluation}

Several evaluation metrics that work on languages with large vocabularies are incompatible with the nature of Toki Pona; for instance, evaluating with analogy performance is impractical due to the fact that Toki Pona's words generally lack parallels. As such, the models are evaluated on two metrics: category centroid proximity with a set of ground truth categories and automated evaluation with agglomerative clustering.

\subsubsection{Category Centroids}

Transferring existing benchmarks such as SimLex-999 \cite{hill2015}, WordSim-353 \cite{finkelstein2001} to Toki Pona is infeasible, due to both the small size of its lexicon and the broad semantic scope of its vocabulary compared to other languages. Instead, this study uses a variation on \citeposs{tsvetkov2015} approach to English model evaluation, creating categories based on WordNet supersenses \cite{miller1995}. While WordNet's original categories do provide semantic groupings for nouns and verbs, adjective supersenses are grouped by syntactic rather than semantic categories; therefore, we based our categories on \citeposs{tsvetkov2014} adjective supersenses in addition to the original noun and verb senses.

Applying English WordNet supersenses directly to Toki Pona may introduce an Anglocentric bias in evaluation \cite{Bender_2011}. To create an effective ground truth for evaluation, categories were adapted to reflect Toki Pona's semantic structure. For instance, because of Toki Pona's small vocabulary, words frequently change parts of speech depending on their syntactic position \cite{lang2014}. As such, distinctions do not exist between categories like \textsc{noun.motion} and \textsc{verb.motion}, and such categories were merged where applicable. Function words like syntactic particles (e.g., \textit{li}, \textit{e}, \textit{pi}) and pronouns (e.g., \textit{mi}, \textit{sina}) were excluded from this analysis to focus on semantic relations. In total, 27 categories were developed over 108 content words, with 125 word-category assignments. For the complete mapping, see Appendix~\ref{sec:categories}.

For each category $C$, a centroid vector $\mu_C$ is calculated by averaging the normalized word vectors $v_w$ for all words $w$ in that category.

\begin{equation}
  \mu_C = \frac{1}{\left|C\right|} \sum_{w \in C} v_w
  \label{eq:C}
\end{equation}

Because words in Toki Pona may belong to multiple domains, we avoid a ``winner-take-all'' measurement of the closest category. Instead, to determine if a word has been successfully categorized, we check whether its cosine similarity to each of its categories' centroids is above a dynamic threshold for each particular assignment. The threshold is determined by the particular word's mean similarity across all categories plus one standard deviation ($\mu + \sigma$). To validate the model's alignment with human judgment, we measure the percentage of words that are successfully retrieved, i.e. had a similarity over the dynamic threshold, by their assigned categories.

\subsubsection{Unsupervised Clustering}

To validate semantic relationships without human-biased semantic categories, agglomerative hierarchical clustering was used to build up clusters based on the locations of words in the vector space. Clusters were combined with the complete linkage criterion; i.e., the distance between two clusters was considered the distance between their two most dissimilar members. This was chosen to favor compact, spherical clusters, and to prevent loosely related words from forming long and incoherent groups that might arise from Toki Pona's polysemy.

Rather than specifying a number of clusters $k$, a hyperparameter sweep was performed over the distance threshold parameter from 0.50 to 0.90, allowing the model to determine the natural number of semantic clusters with varying levels of strictness. The resulting groupings were evaluated with their silhouette scores, selecting the distance thresholds that maximized the value.

Furthermore, a random baseline was established to verify that the resulting silhouette scores' values were statistically significant. A random normal distribution for 108 categorized words was evaluated with the same hierarchical clustering algorithm 30 times.

\subsubsection{Cross-Linguistic Comparison}

We use a method informed by \citeposs{beinborn-choenni-2020-semantic} approach to analyze cross-linguistic drift, using a representational similarity analysis matrix to compare the overall semantic structure of Toki Pona with English. This is done by mapping the Toki Pona categories with their corresponding English translations. For English, we used the pre-trained embeddings trained on the Google News corpus \citep{mikolov2013efficientestimationwordrepresentations}, which contains 300-dimensional vectors for approximately 3 million unique tokens.

\section{Results}

\begin{figure*}
  \centering
  \begin{minipage}[c]{0.8\textwidth}
    \includegraphics[width=\textwidth,alt={Full Corpus model and Pure Toki Pona model full-vocabulary UMAP projections.}]{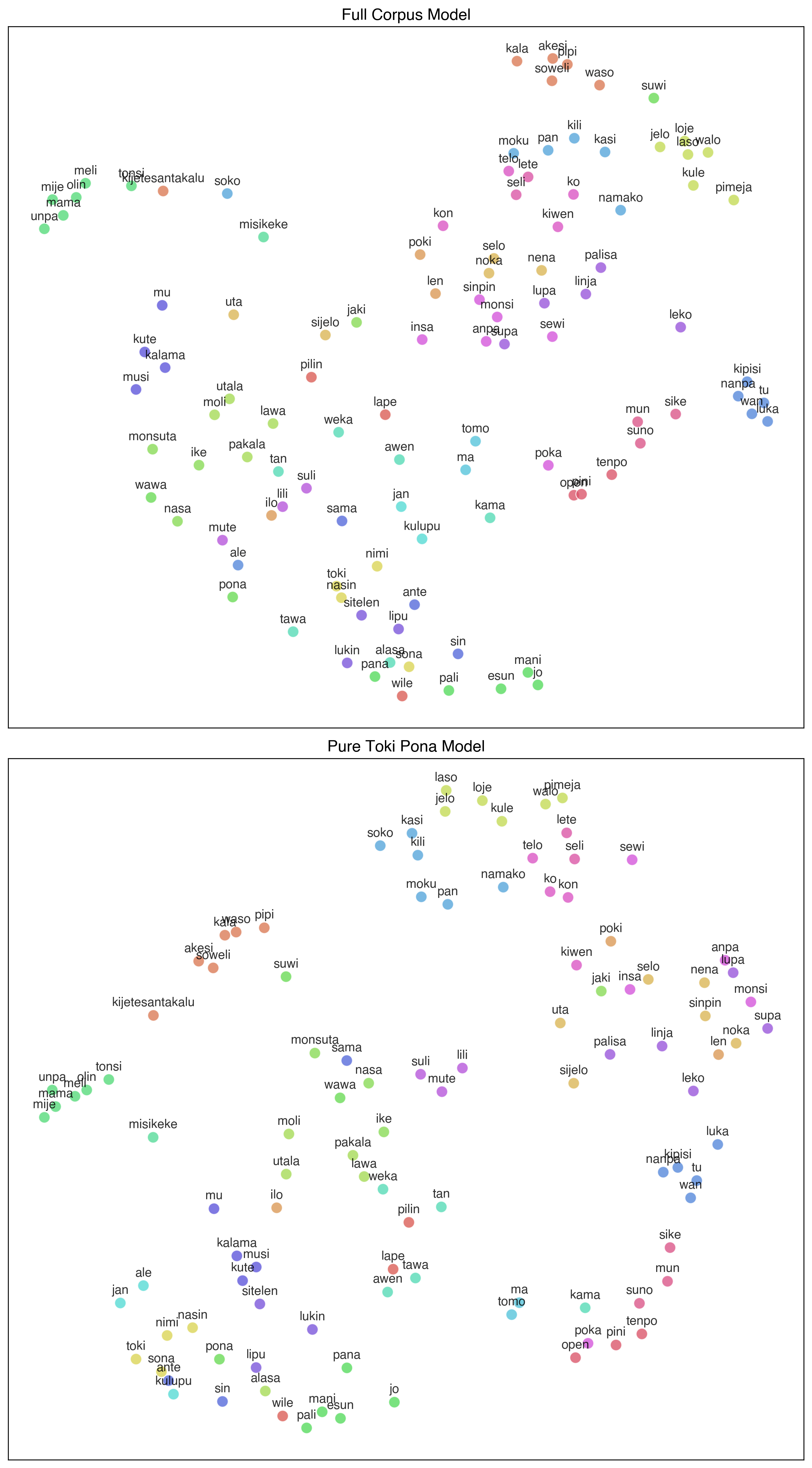}
  \end{minipage}%
  \begin{minipage}[c]{0.2\textwidth}
    \centering
    \includegraphics[width=\textwidth,alt={Nearest category legend for UMAP projections, assigning a color to each category.}]{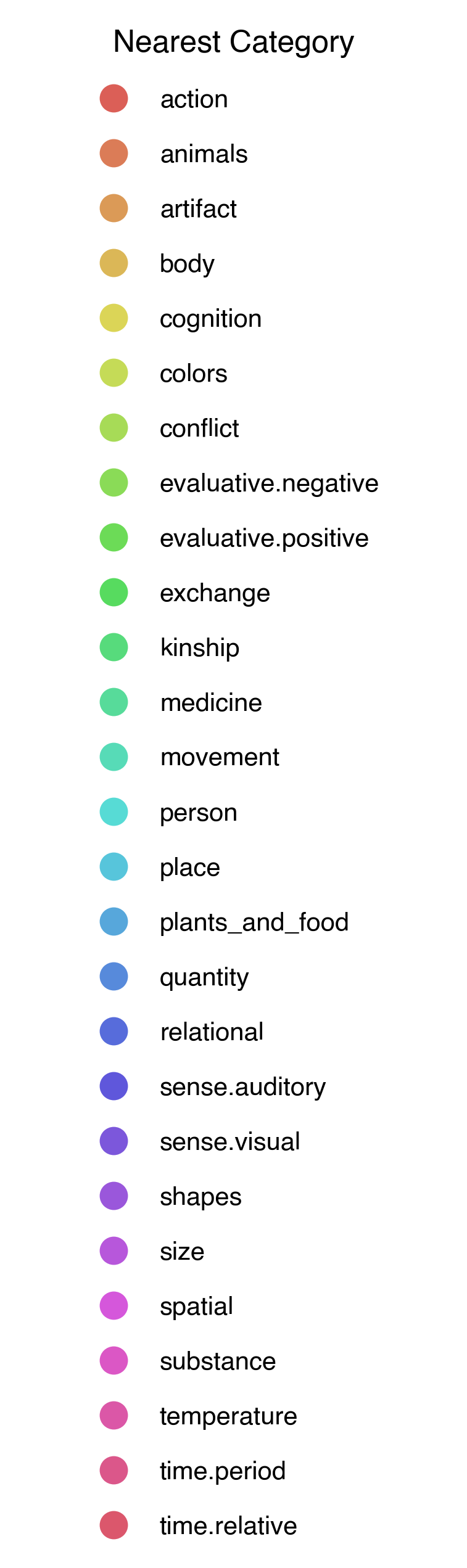}
  \end{minipage}
  \caption{Full-vocabulary semantic space comparison. Models are projected by UMAP with the cosine metric and 15 nearest neighbors. Words are colored by their nearest semantic centroid.}
  \label{fig:global-vis}
\end{figure*}

\subsection{Agglomerative Clustering}

\begin{table}
  \centering
  \small
  \begin{tabular}{lll}
    \toprule
    \textbf{Metric}           & \textbf{Full Corpus} & \textbf{Pure Toki Pona} \\
    \midrule
    Silhouette Score & $0.261$ & $0.267$ \\
    Threshold & $0.38$ & $0.58$ \\
    \bottomrule
  \end{tabular}
  \caption{\label{tab:unsupervised}
    Comparison of clustering metrics between the Full Corpus model and the Pure Toki Pona model.
  }
\end{table}

Both Toki Pona models achieved peak silhouette scores significantly above the random baseline. As shown in Table~\ref{tab:unsupervised}, the baseline had a mean silhouette score of $0.166$ and a standard deviation of $\sigma = 0.0120$; the Full Corpus model's score of $0.261$ and the Pure Toki Pona model's score of $0.267$ are significantly higher than the baseline.

\begin{table*}
  \centering
  \begin{tabular}{llll}
    \toprule
    \textbf{Category} & \textbf{Full Corpus Score} & \textbf{Pure Toki Pona Score} & \textbf{Difference} \\
    \midrule
    \textsc{animals} & $0.864$ & $0.860$ & $-0.004$ \\
    \textsc{colors} & $0.928$ & $0.886$ & $-0.042$ \\
    \textsc{spatial} & $0.899$ & $0.816$ & $-0.083$ \\
    \textsc{evaluative.positive} & $0.858$ & $0.707$ & $-0.151$ \\
    \bottomrule
  \end{tabular}
  \caption{Differences in average cosine similarity of words in specific categories to their category's centroid from the Full Corpus model to the Pure Toki Pona model.}
  \label{tab:filtering-impact}
\end{table*}

The difference in silhouette scores between the Full Corpus model and Pure Toki Pona model is small; the relative stability of the metrics between the two models indicates that the sparse non-Toki Pona data did not have a large impact on the general structure of generated word embeddings. A visualization confirms this stability; Figure~\ref{fig:global-vis} presents a UMAP \cite{mcinnes2018} projection of the vocabulary for both models, colored by their nearest centroid category. The arrangement of clusters and relative locations of words within those clusters indicate that the structure is largely invariant to the presence of occasional non-Toki Pona tokens. For instance, the \textsc{kinship} cluster on the upper right of both projections remains consistently isolated from the remaining vocabulary, along with the \textsc{animals} and \textsc{colors} categories.

Similarly, the jump in the dynamic threshold from 0.38 (Full Corpus model) to 0.58 (Pure Toki Pona model) indicates that in the Pure Toki Pona model, words are more similar to their category centroids on average. This suggests that without the external non-Toki Pona tokens, the minimalist vocabulary of Pure Toki Pona creates a much denser, more interconnected vector space where a higher baseline similarity is required to distinguish a specific category assignment from general semantic noise.

Furthermore, qualitative inspection of the generated clusters demonstrates that both models successfully recover distinct semantic domains without supervision. In both models, \textsc{animals} (e.g., \textit{soweli}, \textit{waso}, \textit{kala}) and \textsc{colors} (e.g., \textit{loje}, \textit{laso}, \textit{jelo}) consistently formed exclusive clusters at optimal distance thresholds. These groupings emerged independently of the manually defined categories used for the centroid analysis, suggesting that the embeddings do truly capture the semantic structure of the language.

While several categories such as \textsc{animals}, \textsc{kinship}, and \textsc{colors} are relatively isolated from the rest of the vocabulary, multiple clusters exhibit significant overlap, notably \textsc{shapes}, \textsc{body}, and \textsc{direction}. This intersection is largely due to Toki Pona's radical polysemy; words for shapes and directions often also double as words for body parts (e.g. \textit{nena} `bump, nose'; \textit{linja} `rope, string, hair'; \textit{sinpin} `front, face'; \textit {selo} `surface, skin'). As such, the embeddings reflect the inherent overlap of these domains in Toki Pona.

\subsection{Category Centroids}

Both Toki Pona models achieved similar retrieval rates with each successfully identifying 124 of the 125 word-category assignments. This corroborates the unsupervised clustering's demonstration that the sets of Toki Pona words are structurally equivalent. However, an analysis of words' proximity to their category's centroid revealed significant differences in density.

\subsubsection{Semantic Density}

The Full Corpus model had a higher mean cosine similarity to category centroids $\mu = 0.879$ compared to the Pure Toki Pona model's $\mu = 0.817$, a difference of $-0.062$. This suggests that non-Toki Pona tokens like names and loanwords are not required to form selected clusters but still provide valuable additional context that pulls related terms closer together in the vector space.

For instance, the named entity \textit{Jutu} `YouTube' is closest to the vectors for \textit{lipu} `website', \textit{ilo} `tool', and \textit{sitelen} `image, video'. The additional data with co-occurrence of \textit{Jutu} and these words is likely to have brought these categories closer together in the vector space, contributing to the increased cluster density.


The impact of removing non-Toki Pona tokens was not uniform across all semantic domains. Table~\ref{tab:filtering-impact} demonstrates that concrete domains like \textsc{animals} tend to be more stable, whereas more abstract categories like \textsc{evaluative.positive} are not as stable.

\begin{figure*}
  \centering
  \pgfplotstableread[col sep=comma]{matrix_labels.csv}{\loadedlabels}

  \begin{tikzpicture}
    \begin{groupplot}[
        group style={
          group size=2 by 1,
          horizontal sep=0.125cm,
          vertical sep=0pt
        },
        width=0.40\textwidth,
        height=0.40\textwidth,
        colormap/viridis,
        point meta min=0,
        point meta max=1,
        colorbar style={
          width=0.3cm,
          ytick={0,0.2,0.4,0.6,0.8,1.0}
        },
        axis on top,
        y dir=reverse,
        xtick={0,...,7},
        ytick={0,...,7},
        enlarge x limits={abs=0.5},
        enlarge y limits={abs=0.5},
        xticklabel style={
          rotate=45,
          anchor=east,
          font=\scriptsize\scshape
        },
        yticklabel style={font=\scriptsize\scshape},
        xticklabels from table={\loadedlabels}{label},
        yticklabels from table={\loadedlabels}{label},
      ]
      \nextgroupplot[title={\textbf{A. Toki Pona (Full Corpus)}}, colorbar=false]
      \addplot[
        matrix plot*,
        mesh/rows=8,
        mesh/cols=8,
        point meta=explicit
      ] table[x=x, y=y, meta=Value, col sep=comma] {matrix_tp.csv};
      \nextgroupplot[
        title={\textbf{B. English (Google News)}},
        yticklabels={},
        colorbar right,
        colorbar style={
          at={(1.05,0.5)},
          anchor=west
        }
      ]
      \addplot[
        matrix plot*,
        mesh/rows=8,
        mesh/cols=8,
        point meta=explicit
      ] table[x=x, y=y, meta=Value, col sep=comma] {matrix_en.csv};
    \end{groupplot}
  \end{tikzpicture}
  \caption{Representational similarity matrices with a subset of Toki Pona categories and corresponding English categories.}
  \label{fig:rsa}
\end{figure*}
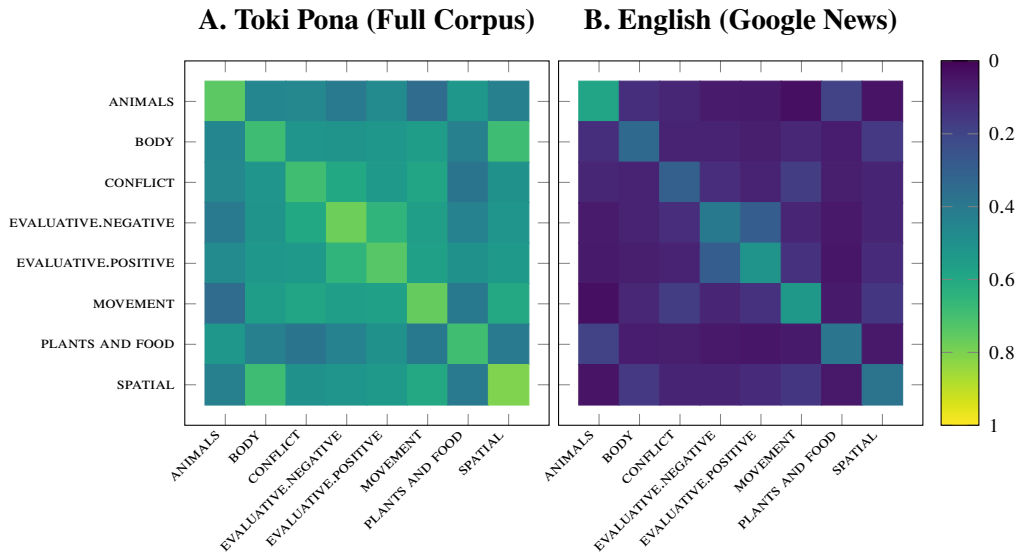

\subsubsection{Stability of Meaning}

Although the density of several categories in the Pure Toki Pona model decreased, the overall structure of the semantic groupings remained largely stable---105 of the 108 words analyzed held the same top category in both models. One of these changes was a shift between two categories---\textit{sinpin} `face, front' moving from \textsc{spatial} to \textsc{body}. Another was a miscategorization of \textit{alasa} `search, try' as a \textsc{movement} word in the Full Corpus model. The third was a miscategorization of \textit{ale} `all, everything' as a \textsc{person} word in the Pure Toki Pona model. This is likely due to the frequent co-occurrence of \textit{ale} with \textit{jan} in the phrase \textit{jan ale} `everyone', as in \textit{toki a, jan ale o} `hello, everyone'.

\subsection{Cross-Linguistic Comparison}

A qualitative analysis of Figure~\ref{fig:rsa} indicates that although the Toki Pona categories are more dense (as expected with the language's radical polysemy along with the lower vector dimensionality), the patterns of relative proximity between categories of words are remarkably consistent across both Toki Pona and English models. For instance, the \textsc{body} and \textsc{spatial} categories are closer together than nearby categories, along with the \textsc{animals} and \textsc{plants and food} categories. For readability, we present Figure~\ref{fig:rsa} with a representative subset of eight categories here; complete matrices for all 27 categories with the same pattern are available in Appendix~\ref{sec:full-rsa}.

\section{Conclusion and Future Directions}

This study investigated whether Word2Vec could generate meaningful embeddings within Toki Pona's highly constrained lexicon of approximately 130 words, while additionally exploring the architectural effects of non-core tokens. Our results indicate that vocabulary size does not constitute a bottleneck for Word2Vec. Even at this extreme lower bound, the model successfully encoded the semantic structure of the language's lexicon and distinguished distinct category clusters without supervision.

Furthermore, a comparison between the Full Corpus and Pure Toki Pona models demonstrates that while core semantic relationships are entirely recoverable in the pure model, the density of the resulting embeddings improves with the inclusion of occasional non-Toki Pona tokens. This suggests that these external elements—primarily proper nouns, loanwords, or instances of code-switching—act as structural bridges within the vector space, effectively refining the positioning of highly polysemous vocabulary. This validity is further supported by our cross-linguistic comparison, which demonstrates that relative distances between semantic categories remain remarkably consistent between Toki Pona and English, confirming that the embeddings capture genuine semantic structures rather than corpus artifacts.

\subsection{Broader Impacts and Future Work}

This work challenges the conventional assumption that fine lexical distinctions and massive vocabularies are prerequisites for effectively encoding meaning via word embeddings. By successfully training a model on a language characterized by extreme polysemy and a minimalist lexicon, we confirm that the distributional hypothesis holds, even at an absolute lower bound.

As a reviewer pointed out, these findings may offer crucial implications for the processing of low-resource and endangered languages in two key ways.

First, many endangered or under-resourced languages possess compact core lexicons or rely heavily on extreme polysemy and compounding (e.g., polysynthetic structures) to convey complex concepts. Our findings provide an empirical proof-of-concept that Word2Vec’s architecture remains robust under these constraints, organizing semantic spaces effectively through distributional density alone.

Second, in real-world, low-resource communities, speakers frequently engage in code-switching or integrate loanwords from dominant regional languages. Our finding that non-core tokens act as a bridge to tighten the vector space without disrupting its underlying structure suggests that NLP pipelines for low-resource languages do not require perfectly curated, "pure" corpora. Instead, the inclusion of code-switched data and lexical borrowings can actively accelerate semantic alignment and stabilize sparse embeddings.

Naturally, these broader computational impacts require empirical validation in natural languages. Future work will focus on applying these insights to natural low-resource, endangered language datasets, specifically investigating how code-switching densities and varying degrees of polysemy affect embedding stability in endangered language documentation.

\section{Acknowledgments}
This study was supported by the President’s Undergraduate Research Travel Award and Professional Development Fund (School of Modern Languages) at the Georgia Institute of Technology, as well as the SCiL 2026 Student Subsidy that helped fund conference travel. The authors are grateful to the reviewers for their constructive feedback and suggestions.

\bibliography{custom}

\appendix

\section{Toki Pona and Corresponding English Categories}
\label{sec:categories}

The full mapping between categories and Toki Pona words used for the centroid distance analysis is detailed in Table~\ref{tab:category-words} along with the English words used for the cross-linguistic comparison.

\begin{table*}[htbp]
  \centering
  \begin{tabularx}{\textwidth}{lXX}
    \toprule
    \textbf{Category} & \textbf{Toki Pona} & \textbf{English} \\
    \midrule
    \textsc{action} & lape, musi, olin, pilin, wile & desire, feel, play, sleep, want \\
    \textsc{animals} & akesi, kala, kijetesantakalu, pipi, soweli, waso & animal, bird, fish, insect, mammal, reptile \\
    \textsc{artifact} & ilo, len, poki & box, cloth, container, machine, tool \\
    \textsc{body} & lawa, luka, monsi, nena, noka, poka, selo, sijelo, sinpin, uta & arm, back, body, face, head, leg, mouth, nose, skin \\
    \textsc{cognition} & lipu, nasin, nimi, sitelen, sona, toki & book, image, knowledge, language, method, way, word \\
    \textsc{colors} & jelo, kule, laso, loje, pimeja, walo & black, blue, green, red, white, yellow \\
    \textsc{conflict} & alasa, moli, monsuta, pakala, utala & break, fear, fight, hunt, kill, monster, war \\
    \textsc{evaluative.negative} & ike, jaki, monsuta, nasa & bad, gross, odd, scary, weak \\
    \textsc{evaluative.positive} & pona, suli, suwi, wawa & awesome, cute, good, important \\
    \textsc{exchange} & esun, jo, mani, pali, pana & buy, give, make, money, sell, work \\
    \textsc{kinship} & jan, kulupu, mama, meli, mije, olin, tonsi, unpa & brother, love, man, parent, sex, sister, woman \\
    \textsc{medicine} & misikeke & cure, doctor, medicine \\
    \textsc{movement} & awen, kama, tan, tawa, weka & come, go, leave, move, stay, wait \\
    \textsc{person} & jan, kulupu & community, group, human, person \\
    \textsc{place} & ma, tomo & country, house, land, room \\
    \textsc{plants and food} & kasi, kili, moku, namako, pan, soko, suwi & bread, food, fruit, mushroom, plant, sugar \\
    \textsc{quantity} & ale, kipisi, luka, mute, nanpa, tu, wan & all, five, many, number, one, part, two \\
    \textsc{relational} & ante, sama, sin & different, new, other, same \\
    \textsc{sense.auditory} & kalama, kute, mu & hear, listen, noise, sound \\
    \textsc{sense.visual} & lipu, lukin, sitelen & image, look, picture, see \\
    \textsc{shapes} & leko, linja, lupa, nena, palisa, sike, supa & block, bump, circle, hole, line, rod, square, surface \\
    \textsc{size} & lili, mute, suli & big, long, short, small \\
    \textsc{spatial} & anpa, insa, monsi, poka, sewi, sinpin & back, bottom, center, front, inside, side, top \\
    \textsc{substance} & kiwen, ko, kon, telo & air, gas, metal, powder, stone, water \\
    \textsc{temperature} & lete, seli & cold, cool, hot, warm \\
    \textsc{time.period} & mun, sike, suno, tenpo & day, era, month, year \\
    \textsc{time.relative} & kama, open, pini, tenpo & finish, future, past, start, time \\
    \bottomrule
  \end{tabularx}
  \caption{Category assignments for Toki Pona and English words.}
  \label{tab:category-words}
\end{table*}

\section{Full Representational Similarity Matrices}
\label{sec:full-rsa}

\begin{figure*} [htbp]
  \centering
  \pgfplotstableread[col sep=comma]{matrix_labels_full.csv}{\fullloadedlabels}
  \begin{tikzpicture}
    \begin{groupplot}[
        group style={
          group size=2 by 1,
          horizontal sep=0.125cm,
          vertical sep=0pt
        },
        width=0.48\textwidth,
        height=0.48\textwidth,
        colormap/viridis,
        point meta min=0,
        point meta max=1,
        colorbar style={
          width=0.3cm,
          ytick={0,0.2,0.4,0.6,0.8,1.0},
          tick label style={font=\tiny}
        },
        axis on top,
        y dir=reverse,
        xtick={0,...,26},
        ytick={0,...,26},
        enlarge x limits={abs=0.5},
        enlarge y limits={abs=0.5},
        xticklabel style={
          rotate=90,
          anchor=east,
          font=\tiny\scshape
        },
        yticklabel style={font=\tiny\scshape},
        xticklabels from table={\fullloadedlabels}{label},
        yticklabels from table={\fullloadedlabels}{label},
      ]
      \nextgroupplot[title={\textbf{A. Toki Pona (Full Corpus)}}, colorbar=false]
      \addplot[
        matrix plot*,
        mesh/rows=27,
        mesh/cols=27,
        point meta=explicit,
      ] table[x=x, y=y, meta=Value, col sep=comma] {matrix_tp_full.csv};
      \nextgroupplot[
        title={\textbf{B. English (Google News)}},
        yticklabels={},
        colorbar right,
        colorbar style={
          at={(1.05,0.5)},
          anchor=west
        }
      ]
      \addplot[
        matrix plot*,
        mesh/rows=27,
        mesh/cols=27,
        point meta=explicit
      ] table[x=x, y=y, meta=Value, col sep=comma] {matrix_en_full.csv};

    \end{groupplot}
  \end{tikzpicture}
  \caption{Complete representational similarity matrices comparing the semantic structure of all 27 defined categories in Toki Pona against English.}
  \label{fig:rsa-full}
\end{figure*}
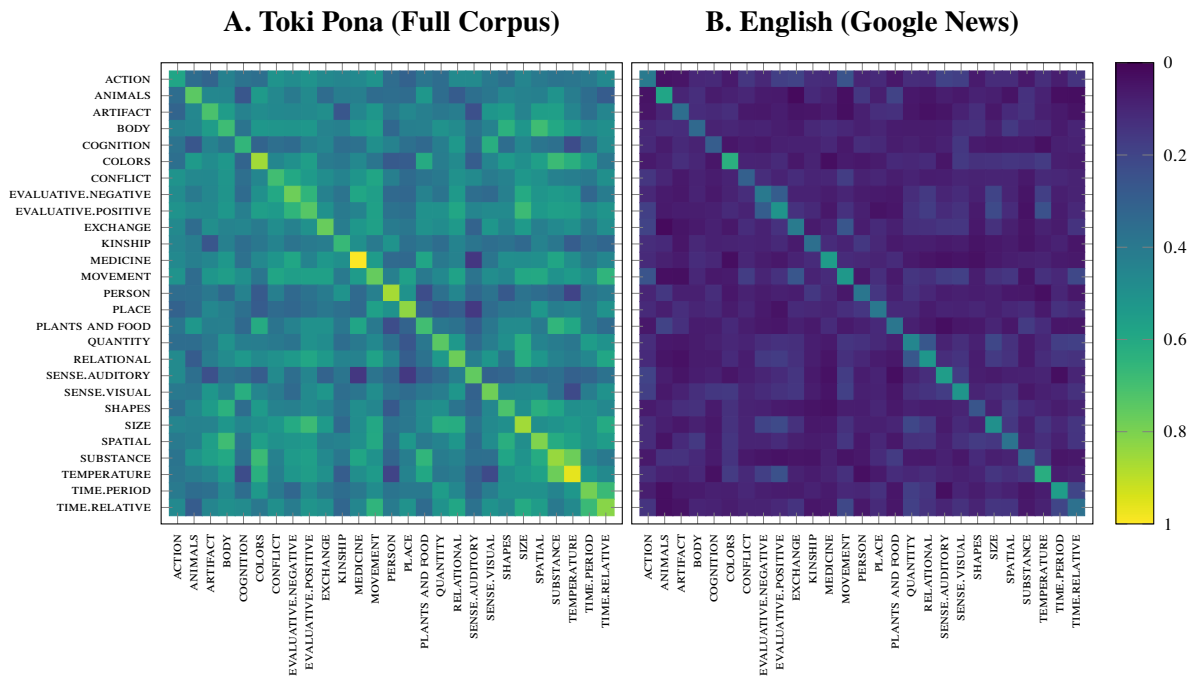

Figure~\ref{fig:rsa-full} presents the complete representational similarity matrices for all 27 categories evaluated.

\section{Code}
\label{sec:code}

All code, trained models, and evaluation data are available at \url{https://doi.org/10.5281/zenodo.20297806}.

\end{document}